%% file: main.tex
\pdfoutput=1
\documentclass[sigconf]{acmart}
\AtBeginDocument{%
 }

\copyrightyear{2024}
\acmYear{2024}
\setcopyright{rightsretained}
\acmConference[CIKM '24] {Proceedings of the 33rd ACM International Conference on Information and Knowledge Management}{October 21--25, 2024}{Boise, ID, USA.}
\acmBooktitle{Proceedings of the 33rd ACM International Conference on Information and Knowledge Management (CIKM '24), October 21--25, 2024, Boise, ID, USA}
\acmDOI{10.1145/3627673.3679732} 
\acmISBN{979-8-4007-0436-9/24/10}

\usepackage{soul}
\usepackage{url}
\usepackage{bm}
\usepackage[utf8]{inputenc}
\usepackage{hyperref}

\usepackage{amssymb}
\usepackage{algorithm}
\usepackage{algorithmic}
\usepackage{amssymb}
\usepackage{subcaption}
\usepackage{subcaption,ragged2e}
\usepackage{appendix}
\usepackage{nameref}
\newcommand{\systemname}{StatioCL} 
\usepackage{balance}
\usepackage{graphicx}

\begin{document}

\title[StatioCL]{\textit{StatioCL}: Contrastive Learning for Time Series via Non-Stationary and Temporal Contrast}

\author{Yu Wu}
\email{yw573@cam.ac.uk}
\orcid{0009-0004-3709-7472}
\affiliation{%
  \institution{University of Cambridge}
  \city{Cambridge}
  \country{United Kingdom}
}

\author{Ting Dang}
\email{ting.dang@unimelb.edu.au}
\orcid{0000-0003-3806-1493}
\affiliation{%
  \institution{University of Melbourne}
  \city{Melbourne}
  \country{Australia}
}

\author{Dimitris Spathis}
\email{dimitrios.spathis}
\email{@nokia-bell-labs.com}
\orcid{0000-0001-9761-951X}
\affiliation{%
  \institution{Nokia Bell Labs}
  \city{Cambridge}
  \country{United Kingdom}
}

\author{Hong Jia}
\email{hong.jia@unimelb.edu.au}
\orcid{0000-0002-6047-4158}
\affiliation{%
  \institution{University of Melbourne}
  \city{Melbourne}
  \country{Australia}
}

\author{Cecilia Mascolo}
\email{cm542@cam.ac.uk}
\orcid{0000-0001-9614-4380}
\affiliation{%
  \institution{University of Cambridge}
  \city{Cambridge}
  \country{United Kingdom}
}

\begin{abstract}
\input{00_abstract}

\end{abstract}



\maketitle

\input{01_introduction}

\input{02_related_work}

\input{03_methods}

\input{04_experiments}

\input{05_conclusion}

\begin{acks}
This work is supported by ERC through Project 833296 (EAR) and Nokia Bell Labs through a donation.
\end{acks}

\bibliographystyle{ACM-Reference-Format}
\vfill\eject
\bibliography{main.bbl}


\end{document}

%% file: 00_abstract.tex
Contrastive learning (CL) has emerged as a promising approach for representation learning in time series data by embedding similar pairs closely while distancing dissimilar ones. However, existing CL methods often introduce false negative pairs (FNPs) by neglecting inherent characteristics and then randomly selecting distinct segments as dissimilar pairs, leading to erroneous representation learning, reduced model performance, and overall inefficiency.
To address these issues, we systematically define and categorize FNPs in time series into \emph{semantic false negative pairs} and \emph{temporal false negative pairs} for the first time: the former arising from overlooking similarities in label categories, which correlates with similarities in non-stationarity and the latter from neglecting temporal proximity. 
Moreover, we introduce \systemname{}, a novel CL framework that captures non-stationarity and temporal dependency to mitigate both FNPs and rectify the inaccuracies in learned representations.
By interpreting and differentiating non-stationary states, which reflect the correlation between trends or temporal dynamics with underlying data patterns, \systemname{} effectively captures the semantic characteristics and eliminates semantic FNPs.
Simultaneously, \systemname{} establishes fine-grained similarity levels based on temporal dependencies to capture varying temporal proximity between segments and to mitigate temporal FNPs. 
Evaluated on real-world benchmark time series classification datasets, \systemname{} demonstrates a substantial improvement over state-of-the-art CL methods, achieving a \textcolor{black}{2.9\%} increase in Recall and a \textcolor{black}{19.2\%} reduction in FNPs. 
Most importantly, \systemname{} also shows enhanced data efficiency and robustness against label scarcity.

%% file: 01_introduction.tex
\section{Introduction}
Time series data play a crucial role across various fields, such as finance, climate modeling, and healthcare~\cite{bagnall2018uea,ruiz2021,tang_finance, Ismail_Fawaz_2019, wu2023udamaunsuperviseddomainadaptation}. 
While deep neural networks, including transformers, have significantly improved time series modeling, these methods often rely on supervised learning and high-quality annotations~\cite{zhang2023survey}.
However, labeling these data is often challenging and time-consuming~\cite{deldari2022}.
For example, in fitness monitoring, it can be difficult to manually interpret Inertial Measurement Unit (IMU) sensory data from wearable devices, often resulting in ineffective labeling of the data~\cite{tang2021}. Therefore, the scarcity of adequate and reliable labels for time series data poses obstacles to effective time series modeling and analysis. 
With vast amounts of unlabeled data being generated by various devices, there has been a growing interest in efficiently modeling time series in an unsupervised manner for time series classification tasks~\cite{franceschi2020unsupervised, Audibert2020}. 

Recently, self-supervision has become a leading method in unsupervised learning, with Contrastive Learning (CL) being one of its most prominent techniques.
CL is designed to capture the inherent characteristics of data without requiring manual labeling and has proven effective across various fields~\cite{arora2019theoretical, moco, saeed2020contrastivelearninggeneralpurposeaudio,Yang_2023}. Its primary goal is to create an embedding space where similar (positive) pairs are attracted while dissimilar (negative) pairs are repelled~\cite{huynh2022boosting}. Although CL has shown success in vision~\cite{Verma2021,robinson2021contrastive,khosla2021supervised,park2022fair}, text~\cite{pan2021contrastive, wei2022semisupervised,liu2014contrastive}, and speech~\cite{baevski2020wav2vec,ma2021contrastive,chung2021w2vbert} domains, its application to time series remains limited. Current CL methods for time series data mainly adapt data augmentation techniques from vision-related approaches~\cite{chen2020simple,ts2vec}, or employ domain-specific knowledge~\cite{zhang2022} and temporal pattern understanding~\cite{tonekaboni2021unsupervised,eldele2021} to enhance representation learning.
However, these approaches often fail to comprehensively capture the intrinsic similarities and characteristics within time series data, leading to sub-optimal representations and limited performance and data efficiency in downstream tasks.

Notably, current methods generally adhere to the principle~\cite{chen2020simple}, \textit{i.e.}, creating positive pairs from augmented versions of the same sample and negative pairs from randomly selected, distinct samples. While positive pairs created in this way are generally reliable due to their generation from the same sample sharing similar semantic content, the random selection of negative pairs often leads to \emph{false negative pairs} (FNPs). FNPs occur when randomly chosen negative pairs share the same label categories, which can hinder the effectiveness of the learned representation~\cite{huynh2022boosting}. 
Furthermore, in the context of time series data, this mechanism also overlooks the similarities inherent in time series signal patterns, particularly in cases where temporally proximate segments exhibit similar characteristics. 
For example, in a dataset monitoring heart rates, segments of data recorded within short intervals might display similar patterns due to the consistent nature of heart rhythms over these periods. 

\begin{figure}
    \centering
    \includegraphics[width=0.4\textwidth]{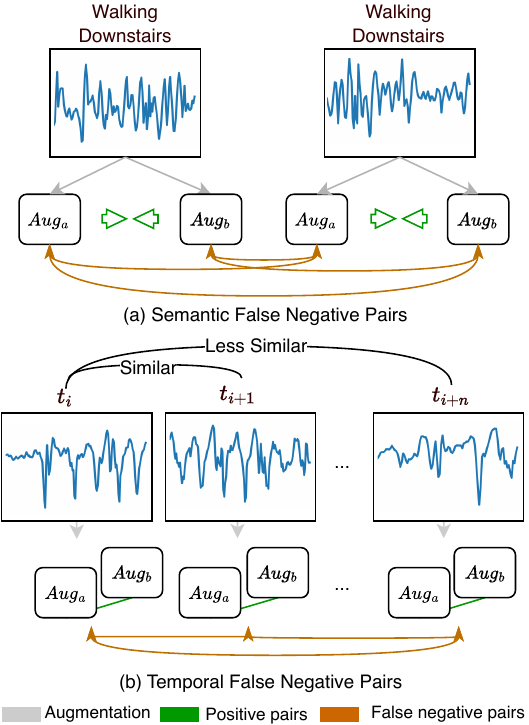}
    \caption{ 
    Identification of False Negative Pairs in Time Series Contrastive Learning Due to Random Selection: (a) \textbf{Semantic FNPs}, occurring when the selection process ignores the similarity between labels and (b) \textbf{Temporal FNPs}, resulting from neglecting the similarity in terms of temporal proximity. 
    }
    \label{fig:fnp}
\end{figure}

Therefore, effectively avoiding these FNPs is crucial for developing time series representation models. Firstly, failing to recognize them as similar segments and arbitrarily labeling them as negative pairs may mislead models, leading to suboptimal performance in learning common patterns~\cite{huynh2022boosting, Sun2023LearningAS}. This, in turn, adversely affects the performance of downstream time series classification. 
Furthermore, eliminating FNPs may decrease the model's confusion and errors during pre-training. Given that representative information is captured during the pre-training phase, a reduced amount of data is required to fine-tune the model for downstream tasks. This aspect assumes greater significance in light of the scarcity of labeled data in real-world scenarios. Consequently, the elimination of FNPs can contribute not only to improved accuracy but also to increased efficiency.

As such, we first identify two distinct types of FNPs to highlight these oversights based on non-stationarity, which are correlated with semantic patterns, and temporal dependencies, both of which are particularly pertinent to the unique nature of time series.
The first type, termed \emph{semantic false negative pairs}, occurs when pairs from the same class are mistakenly identified as dissimilar. This issue is prevalent across various domains, including time series data. For instance, in human activity recognition, two segments of IMU data representing the same activity of 'walking' might erroneously be considered a negative pair in CL, as shown in Figure~\ref{fig:fnp}(a). The second type, termed \emph{temporal false negative pairs}, is unique to time series data.
These are pairs that exhibit similar patterns and are temporally proximate. Traditional CL methods might neglect this temporal continuity and incorrectly treat adjacent segments as negative pairs, as shown in Figure~\ref{fig:fnp}(b). 
Avoiding these FNPs in time series requires a nuanced understanding of both semantic content and temporal structure. While recent approaches have attempted to address this challenge by using hierarchical clustering to refine pair construction, they often face limitations in terms of efficiency and their ability to comprehensively capture the intrinsic characteristics of time series data~\cite{mhccl}.
 
This paper proposes a novel contrastive learning framework named \textit{\systemname{}}, specifically tailored to mitigate both semantic and temporal FNPs. 
By effectively addressing the challenges of misleading model training caused by FNPs in time series data, \systemname{} aims to improve the performance of time series classification tasks, where accurately capturing the semantic and temporal characteristics of the data is crucial for learning the common patterns among time series segments.
Specifically, \systemname{} employs two key strategies that harness the intrinsic characteristics of time series data: non-stationarity and temporal dependency. Non-stationarity refers to the phenomenon where the statistical properties of time series segments change in non-linear ways, often correlating with underlying patterns. For example, in epilepsy classification, seizure segments show strong non-stationarity, while non-seizure segments are relatively stationary. By performing a statistical test to determine stationarity before training, we can glean crucial semantic insights to help construct more accurate negative pairs. \systemname{} leverages this and proposes the non-stationary contrast approach by identifying pairs with differing non-stationarity states as negative to alleviate \emph{semantic false negative pairs}.
Additionally, to alleviate the \emph{temporal false negative pairs}, \systemname{} introduces a temporal module that prioritizes temporal proximity as a key factor in evaluating the similarity between segments. We propose a method to reweight negative pairs based on their temporal distance from the anchor sample, assigning lower weights to closer segments. More specifically, the reweighting mechanism is applied within a defined proximity range, allowing for a more refined approach to negative pair selection. Our contributions are summarized as follows:
\begin{itemize}
    \item We systematically define and differentiate, for the first time, two types of false negative pairs in time series data: semantic and temporal false negative pairs.

    \item Recognizing the significant correlation between non-stationarity and temporal dependencies with regard to underlying semantics, we develop two distinct strategies within \systemname{} to reduce FNPs and enhance representation learning.

    \item We conduct extensive experiments to evaluate \systemname{} on 34 real-world time series benchmark datasets, covering various time series modalities, including sensory signals, biosignals, and general temporal data. The results showcase an average reduction in FNPs by 19.2\% and a 2.9\% relative improvement in Recall for downstream classification tasks.
    
    \item Further analyses reveal intriguing properties of \systemname{} concerning data efficiency and label scarcity robustness. Notably, when only 10\% of fine tuning data is used, \systemname{} demonstrates an average improvement of \textcolor{black}{3.1\%} in accuracy over other contrastive learning methods and a \textcolor{black}{5.4\%} improvement over traditional supervised methods across all datasets. 
\end{itemize}

%% file: 02_related_work.tex
\section{Related Work}
\subsection{Contrastive Learning for Time Series Data}
By learning from unlabeled data, contrastive learning can alleviate the need for extensive labeled datasets. This is particularly beneficial for time series classification, where labeled data may be scarce or difficult to acquire.
Current CL methodologies for time series data can be broadly categorized into three primary streams. 
The first stream concentrates on developing advanced data augmentation to improve model robustness~\cite{ts2vec,zhang2022,yang2023simper,shi2021DTW,luo2023time}. Although these augmentation techniques help generate diverse and informative positive pairs, they neglect the intrinsic characteristics of time series data, and some rely on meta-learning, which reduces the method's efficiency.
Another set of strategies enhances the structure of contrastive pairs beyond the segment level, incorporating subject-level considerations by leveraging inter-subject variability or considering trial-level consistency through neighboring and non-neighboring samples~\cite{intrainter,clocs,wang2023contrast}. 
Nevertheless, these two categories are mostly inherent in the vision-related pairs construction approach ~\cite{chen2020simple} and tend to ignore the label or temporal semantics that might lead to false negative pairs. 
A third stream exploits the intrinsic in time series data, such as temporal dependencies, to enhance time series representation learning.
For instance, CPC and its advanced method of Temporal and Contextual Contrasting (TS-TCC) aim to learn robust representation by predicting the future states based on the past states~\cite{eldele2021,oord2019}.
Temporal Neighborhood Coding (TNC) also leverages the temporal dependencies with the assumption that close segments (neighbor) are similar while far away (non-neighbor) are dissimilar to design contrastive models~\cite{tonekaboni2021unsupervised} and contrasting neighboring and non-neighboring samples.
However, these approaches often neglect finer-grained temporal continuity by arbitrarily categorizing close and distant segments.
As a result, current methods generally overlook label semantics or fine-grained temporal dependencies inherent in the data and generate unreliable pairs, \textit{i.e.}. These false negative pairs lead to potential inaccuracies for downstream time series classification.

\subsection{False Negative Pairs in SSL}
Very few studies have focused on addressing the challenges of FNPs. One category of methods~\cite{byol,caron2021unsupervised} only infers positive pairs through both the student and teacher models and does not contrast against negative samples.
Most of these advancements are primarily focused on the image domain~\cite{huynh2022boosting, wu2020conditional, Sun2023LearningAS}.
In the context of time series data, one recent approach, MHCCL, targets \emph{semantic FNPs} by employing a hierarchical clustering method and subsequently using the cluster indices as pseudo labels to mitigate the incidence of FNPs. However, the mechanism focuses solely on semantic FNPs and significantly increases computation complexity during training, making it unsuitable for real-time processing in practical applications. 
The unique challenges of temporal dynamics that lead to the \emph{temporal FNPs} presented by time series data have received comparatively less attention. 
Although TNC can, to some extent, alleviate the \emph{temporal false negative pair} problem by contrasting negative samples with non-neighbors positioned far from the current segment, it still relies on random selection and treats these as absolute negative pairs, which neglects the fine-grained proximity and label semantics across the time axis. 

In contrast, our work seeks to fill this gap by reducing the occurrence of both types of false negative pairs by leveraging non-stationarity and temporal dependencies inherent in time series data. Therefore, \systemname{} can navigate the intricacies of time series data and achieve improved performance and efficiency.

%% file: 03_methods.tex
\section{Methods}
\begin{figure*}
    \centering
    \includegraphics[width=\textwidth]{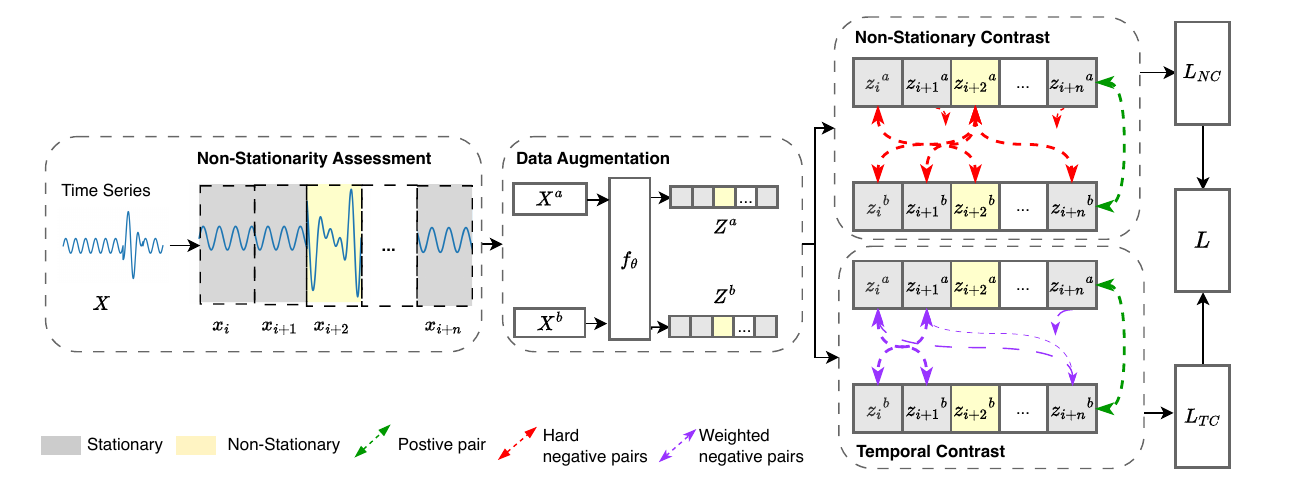}
    \caption{\textbf{\systemname{} Framework Overview:}
    Input sequence $X$ is first passed through  \textbf{(1) Non-stationarity Assessment} module to get the non-stationary state for each segment. \textbf{(2) Augmentation} module will then generate two augmentation views and encode augmented views into feature space. After that, each view will be passed to \textbf{(3) Non-stationary Contrast} module to construct negative pairs based on non-stationary states and reduce semantic FNPs. \textbf{(4) Temporal Contrast} module to create weighted negative pairs based on time differences and alleviate temporal FNPs. Finally, the overall loss $L$ is calculated by combining $L_{NC}$ and $L_{TC}$. 
        }
    \label{fig:framework} 
\end{figure*}

\subsection{Problem Definition}
Given an unlabeled \( N \) multivariate time series denoted by \( \mathbf{X} = \{\mathbf{x}_i\}_{i=1}^{N} \), each time series \( \mathbf{x}_i \in \mathbb{R}^{T \times V} \) encapsulates a sequence of observations across \( V \) 
distinct variables recorded at \( T \) discrete time points.
We aim to design a mapping function, $f_{\theta}$, parameterized by a deep neural network and trained in a novel self-supervised contrastive scheme to learn effective time series representations. 
Specifically, this scheme is distinctive due to its enhanced pair construction mechanism to address the unique challenges of false negative pairs in time series data, such as disregarding label semantics or temporal dependencies. 
The goal is to transform time series segments into informative representations that encapsulate inherent patterns and characteristics, thereby alleviating the challenge and facilitating more effective and efficient application in various downstream tasks. 

\subsection{Non-Stationary and Temporal Contrast} 
\subsubsection{\textbf{Overview.}}
The overall contrastive learning architecture of \systemname{} is shown in Figure~\ref{fig:framework}.  
For each \textit{anchor}, input segment \( \mathbf{x}_i \), its non-stationary state $l_i$ is labeled by the non-stationarity assessment module. 
Then, the data augmentation module will generate two augmented views using weak and strong augmentations. 
These augmented time series are then transformed into the latent space by the model, $f_{\theta}$, to obtain the corresponding representations $ {\bm z_i}^a = f_{\theta}({\bm x_i}^a)$ and ${\bm z_i}^b = f_{\theta}({\bm x_i}^b)$, respectively. 

To mitigate the generation of false negative pairs in traditional CL methods, in the pair construction module, \systemname{} employs two mechanisms that utilize the non-stationarity and temporal dependencies intrinsic within time series data to improve the quality of negative pair selection and reduce the occurrence of false negatives. The details for each module are presented as follows.

\subsubsection{\textbf{Non-Stationarity Assessment.}}
\begin{figure}[h!]
  \begin{subfigure}[b]{0.45\linewidth}
    \includegraphics[width=\linewidth]{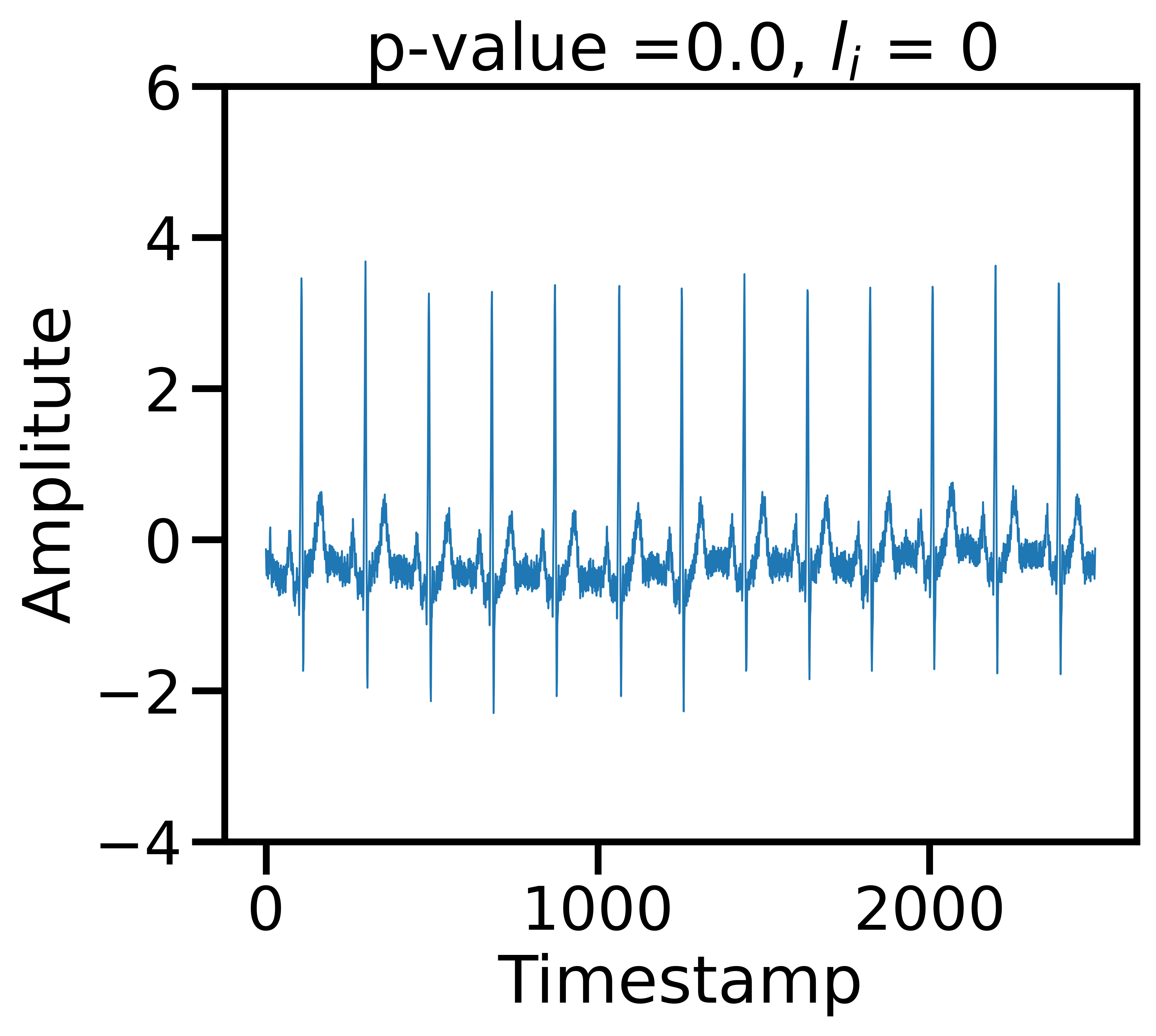}
    \caption{Stationary ECG}
  \end{subfigure}
  \begin{subfigure}[b]{0.45\linewidth}
    \includegraphics[width=\linewidth]{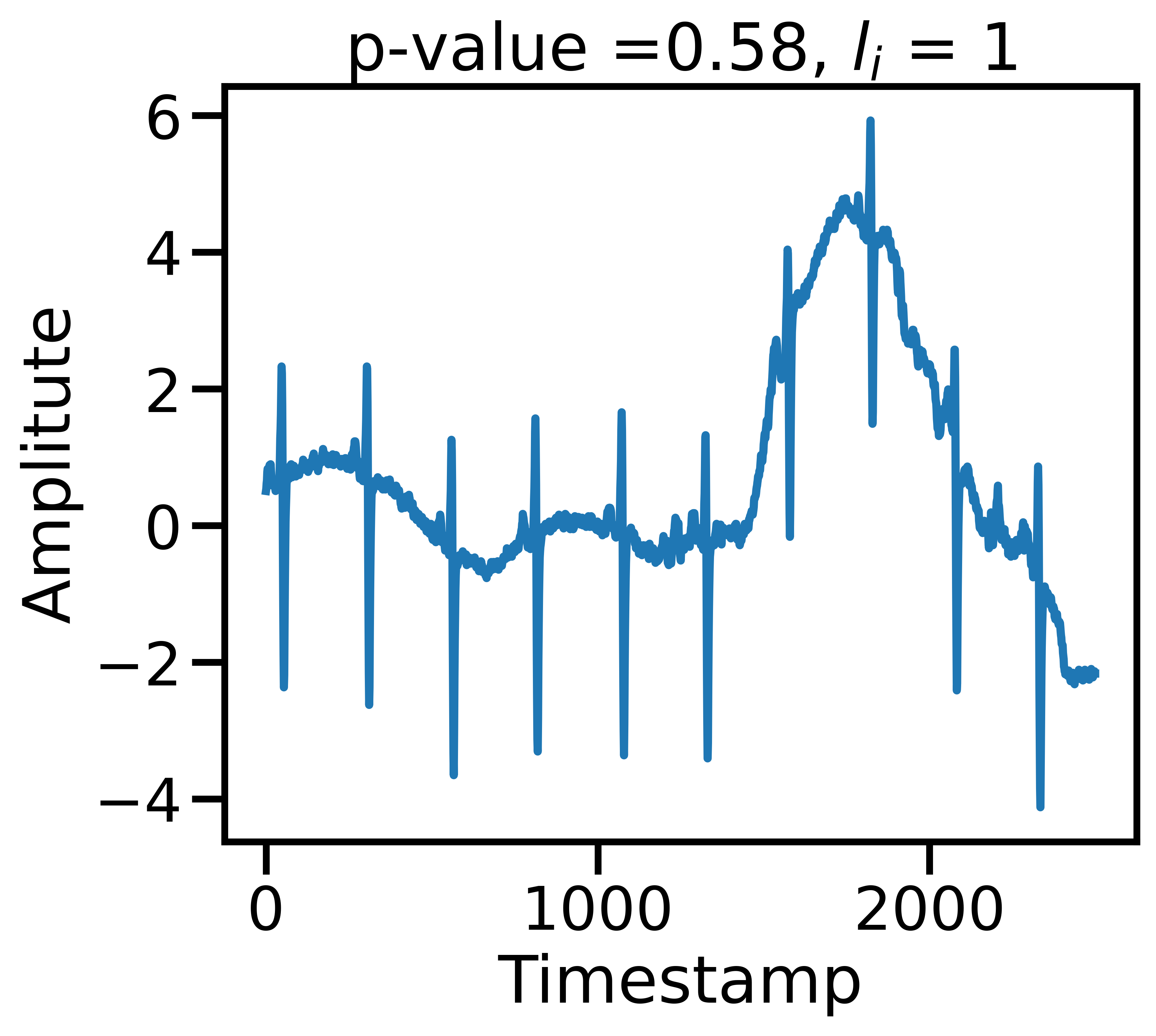}
    \caption{Non-stationary ECG}
  \end{subfigure}
  \caption{\textbf{Example of Stationary and Non-Stationary ECG Signal}.
  If p-value $<$ 0.01, the stationary state $l_i$ is set to 0; otherwise,  $l_i$ is assigned a value of 1.
  } 
  \label{fig:fig3}
\end{figure}

To evaluate the non-stationary state of raw time series data, we employ the Augmented Dickey-Fuller (ADF) statistical test, a widely used method for detecting unit roots in time series data~\cite{mushtaq2011augmented}. Applying this test, we compute a p-value for each segment \( \mathbf{x}_i \) and set a threshold to distinguish between stationary and non-stationary states $l_i$. 
A p-value exceeding the predefined threshold suggests the data's non-stationarity, leading us to label such segments as $l_i = 1$(non-stationary) in our framework. Conversely, segments with p-values below the threshold are deemed stationary and labeled as 0. 
This approach is exemplified in Figure~\ref{fig:fig3}, using ECG signal data, showcasing the effectiveness of our threshold-based classification in identifying \emph{semantic false negative pairs}.

\subsubsection{\textbf{Data Augmentation.}}Data augmentation is achieved by the combination of weak (jitter-and-scale) and strong (permutation-and-jitter) strategies following TS-TCC~\cite{eldele2021}. The aim is to enhance the generalization capability of the system across different types of time series.

\subsection{False Negative Pairs Elimination}
Similar to other contrastive methods, we first define positive pairs as different augmentations of the same input. Specifically, for each latent representation ${\bm z_i}$, its augmented views ${\bm z_i}^a$\ and ${\bm z_i}^b$ maintaining the same stationarity state serves as the positive pair.
Following that, we use two different methods to build negative pairs in the pair construction module to alleviate both semantic and temporal false negative pairs. 

\subsubsection{\textbf{Non-Stationary Contrast Module.}}
The non-stationary contrast module aims to address \emph{semantic false negative pairs}. Our work constructs enhanced negative pairs by leveraging the underlying correlation between non-stationarity and label semantics in time series data. By pushing representations with distinct non-stationary characteristics away in the latent space and using a specially designed loss function, semantic FNPs can be effectively eliminated, leading to improved representations.

Specifically, this module introduces non-stationarity as an inductive bias for negative pair selection. Utilizing the Augmented Dickey-Fuller (ADF) test results as prior knowledge, we assign a binary non-stationary label, $l_i$, to each segment $\bm x_i$; where $l_i = 1$ indicates a non-stationary state and $l_i = 0$ indicates a stationary state. 

For the construction of contrastive negative pairs, segments with differing stationarity states are specifically chosen to form negative pairs, i.e., pairing $\bm z_i^a$ with $\bm z_j^b$ where $i \neq j$ and $l_i \neq l_j$. Such pairs are designated as \textit{hard-negative pairs}.
This approach ensures robust negative pair selection, enhancing the model's ability to discriminate between distinct stationarity states and, by extension, different underlying classes.
To reinforce this learning process, we introduce a tailored contrastive loss function. For a given sample $x_i$, the loss is computed as:
\begin{equation}
    {L_{NC}}^{i} = -log(\frac{exp(sim({\bm z_i}^a, {\bm z_i}^b)/\tau)}{{\sum_{j=1 [i\neq j,l_i \neq l_j]}^{2N}}exp(sim({\bm z_i}^a, {\bm z_j}^b))/\tau})) 
    \label{eq:eq1}
\end{equation}
\noindent where $N$ denotes the batch size and $\tau$ is the temperature constant.

This customized approach effectively minimizes and maximizes the similarity between positive and hard-negative pairs respectively, enabling \systemname{} to capture the critical relationship between non-stationarity and class distinctions, thereby eliminating \emph{semantic false negatives pairs} and enhancing performance in downstream classification tasks. 

\begin{figure}
    \centering
    \includegraphics[width=0.4\textwidth]{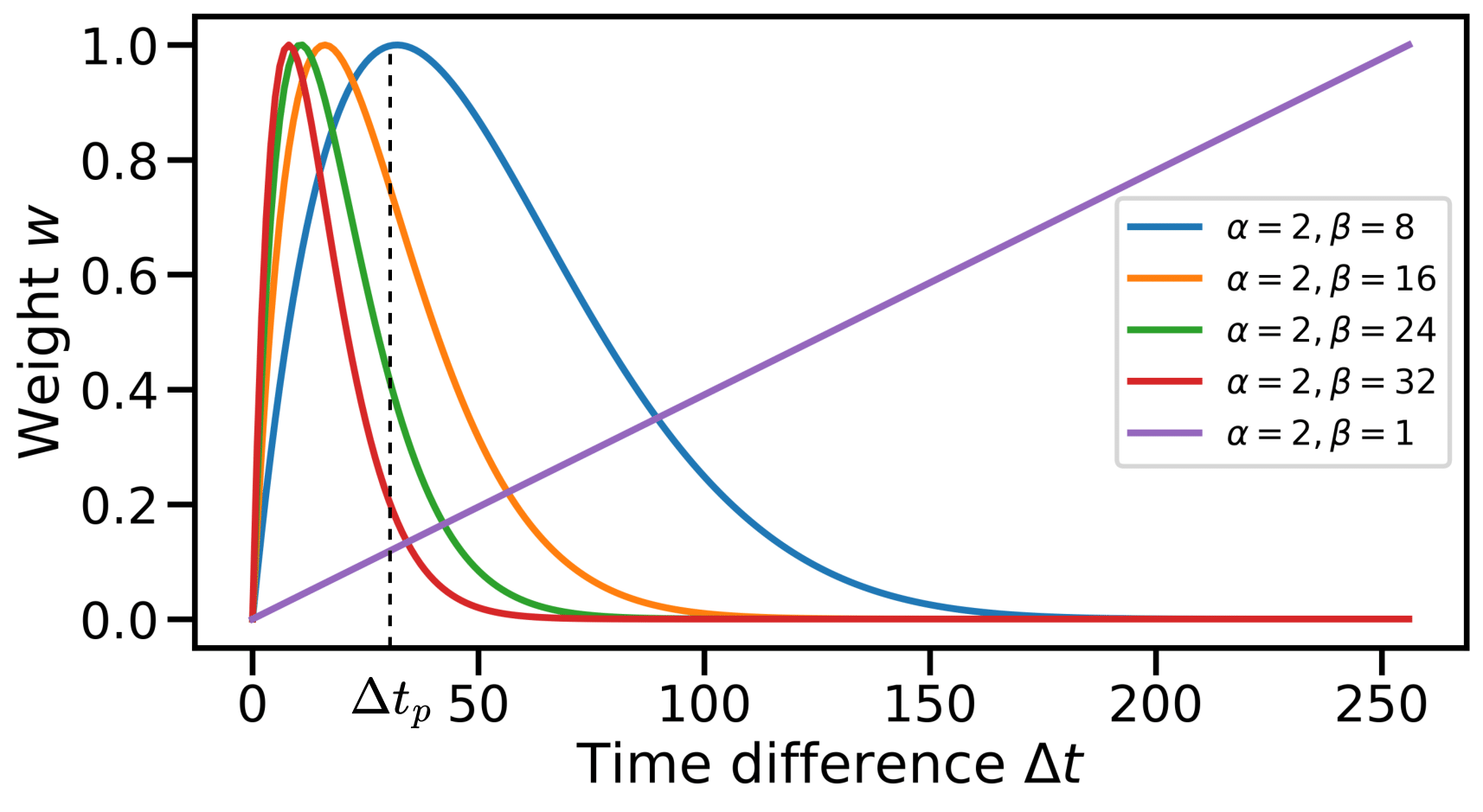}
    \caption{\textbf{Weight Distribution for Soft-Negative Pairs:} The histogram depicts the assigned weights to negative pairs based on their time difference, following the beta distribution.}
    \label{fig:fig4}
\end{figure}

\subsubsection{\textbf{Temporal Contrast module.}}
In \systemname{}, the temporal contrast module improves the construction of
negative pairs, which we term \emph{soft-negative pairs}, to alleviate \emph{temporal false negative temporal pairs} based on the refined assumption utilizing temporal dependency. 
The refined assumption is that adjacent segments in time series data exhibit strong temporal dependencies over a certain period, and as this dependency diminishes over time, the similarity and relevance between these segments decrease. Then, we introduce a weighted mechanism to capture this temporal dependency by considering the correlation between similarity and temporal proximity among negative pairs.

Specifically, these weights are parameterized using a Beta distribution to determine the degree of similarity between each latent representation $\bm z_i^a$ and another representation $\bm z_k^b$ based on their time difference, $\Delta t = \left|i-k\right|$, as illustrated in Figure~\ref{fig:fig4}. This approach is grounded in the premise that segments closer in time often exhibit higher temporal dependencies. Consequently, as the time difference increases, the segments become less similar and are more likely to form negative pairs. Initially, an increasing trend in weights is observed with respect to increasing temporal proximity. However, over a longer time span, it may become ambiguous to determine temporal dependencies. Instead of completely ignoring pairs with a longer time span, we account for scenarios where certain patterns, such as heart rate, may exhibit similar trends even over extended periods, thus making them less likely to be negative pairs. This consideration results in assigning lower weights to such pairs, leading to a decreasing trend in the latter half of the Beta distribution.'

More importantly, the Beta distribution offers significant flexibility in capturing these soft negative pairs. With two parameters, ${\alpha}$ and ${\beta}$, controlling the weight distributions, which are learnable parameters, we can effectively approximate and optimize the temporal dynamics for different data types and applications in constructing negative pairs. For instance, with ${\alpha = 2}$ and ${\beta = 8}$ (as shown by the blue curve in Figure~\ref{fig:fig4}), the weight $w_i$ initially increases as $\Delta t$ grows, reflecting decreasing similarity. It reaches a maximum value of 1 at an optimized time range ($\Delta t_p$). As $\Delta t$ extends beyond this range, reflecting diminishing long-term temporal dependency, the influence of $w_i$ also decreases, following a trend that converges to zero. By varying ${\alpha}$ and ${\beta}$, we can learn different weight parameters, \textit{i.e.}, fast or slow temporal dynamics in the slope of the curves, as well as long or short temporal dependencies indicated by larger or smaller $\Delta t_p$ values. Additionally, different combinations of ${\alpha}$ and ${\beta}$ can result in linear and exponential weight distributions, thereby relaxing constraints in specific weight distribution shifts and allowing the model to learn from the data itself.

Finally, we formulate a comprehensive weighted loss function that fully enhances the aspect of temporal consistency and pair constructions. Similarly as Equation~\eqref{eq:eq1}, for an input sample $\bm x_i$
with timestamp index $i$,
we have the contrastive loss as follows:
\begin{equation}
    {L_{TC}}^{i} = -log(\frac{exp(sim({\bm z_i}^a, {\bm z_i}^b)/\tau)}{{\sum_{k=1 [i\neq k,l_i = l_k]}^{2N}}\bm w_i * exp(sim({\bm z_i}^a, {\bm z_k}^b))/\tau}))
\end{equation}
\begin{equation}
    w_i = \frac{\left | t_i - t_k \right |^{\alpha-1}(1-\left | t_i - t_k \right |)^{\beta-1}}{B(\alpha, \beta)}
\end{equation}

\begin{equation}
    B(\alpha, \beta) = \frac{\Gamma(\alpha)\Gamma(\beta)}{\Gamma(\alpha + \beta)}
\end{equation}

\noindent where $\Gamma$ is to ensure distribution is normalized. As such, the proposed mechanism can enhance the similarity of proximate close pairs while distancing samples farther apart in a more soft and dynamic manner. 

\subsection{Overall Loss}
The final fine-grained contrastive loss is derived by combining the non-stationary and temporal dependency loss. This combined loss
captures the signal characteristics by defining different degrees of similarity within the time series. The overall loss is defined as:
\begin{equation}
    L = \sum_i[\lambda 
 {L_{NC}}^{i} + (1 - \lambda) {L_{TC}}^{i}]
\end{equation}
where $\lambda$ is the scalar hyperparameter denoting the relative weight of each loss.

Therefore, with the formulation of the final loss, our approach, \systemname{}, can significantly alleviate the impact of both semantic and temporal false negative pairs. This alleviation is pivotal in creating a cleaner and more accurate representation space.

%% file: 04_experiments.tex
\section{Experiments}
\subsection{Experiment Settings}

\subsubsection{\textbf{Datasets.}} 
\input{04_experiments_settings}

\subsubsection{\textbf{Implementation Details.}} 
During pretraining, we employ Convolutional Neural Networks (CNN) as encoders due to their generalized capability to effectively capture distinctive features. Specifically, the network architecture includes a 3-layer 1-D CNN followed by a fully connected layer with an output dimension of 64.
To determine the appropriate levels of similarity in non-stationarity, we optimize the ADF threshold using a grid search, with values ranging from 0.0001 to 0.05, depending on the specific characteristics of each signal. For example, the IMU signal for HAR is sensitive to small body movements, resulting in noisy time series that affect the ADF test statistics. Therefore, we use a high threshold to guarantee a reasonable $l_i$ from the ADF test results.
For capturing temporal dependencies using the beta distribution, we empirically select $\alpha=2$ and fine tune the $\beta$ value through a grid search over the set {8, 16, 24, 32}. The choice of beta values depends on the nature of the data, such as whether it exhibits fast-changing temporal dynamics with short-term correlations (lower beta values) or slow-changing dynamics with long-term correlations (higher beta values). The optimized beta values are 8 for the HAR, Sleep-EDF, and Epilepsy datasets and 24 for the ECG Waveform dataset. 
To control the relative weight of each loss term, we optimized $\lambda=0.5$ for the HAR, Epilepsy, and Sleep-EDF datasets and $\lambda=0.1$ for the ECG dataset.

Regarding training specifics, we set the batch size to 256 for the ECG dataset and 128 for the others due to their smaller size. We use the Adam optimizer with a learning rate of 3e-4, weight decay of 3e-4, and moment decay rates $\beta_1=0.9$ and $\beta_2=0.99$. Each model is trained for 150 epochs.
To ensure the reliability \footnote{https://github.com/yvonneywu/statiocl} of our results, we repeat all experiments 5 times with different random seeds and report the mean and standard deviation of the performance metrics.
Finally, all models are implemented using PyTorch, and the experimental evaluations are conducted on NVIDIA A100-SXM-80GB GPUs. 


\begin{table*}
  \centering 
  \small 
  \caption{\textbf{Comparative Performance on Time Series Classification:} This table illustrates the accuracy (ACC) and macro-average F1 (MF1) scores achieved by various methods across different datasets. \textbf{Bold} indicates the best performance for each metric, while \underline{underscore} represents the second-best performance.}
  \begin{tabular*}{\textwidth}{ @{\extracolsep{\fill}} lllllllllll}
  \toprule
              & \multicolumn{2}{c}{HAR} & \multicolumn{2}{c}{Sleep-EDF} & \multicolumn{2}{c}{Epilepsy} & \multicolumn{2}{c}{PenDigits} & \multicolumn{2}{c}{ECG Waveform} \\
    \midrule
Methods       & ACC        & MF1        & ACC           & MF1           & ACC           & MF1    & ACC           & MF1      & ACC          & AUPRC \\ \midrule
Supervised          & 90.1 $\pm$ 2.5 & 90.31 $\pm$ 2.24 & 83.4 $\pm$ 1.4 & 74.78 $\pm$ 0.86 & 96.7 $\pm$ 0.2 & 94.52 $\pm$ 0.43 & 97.5 $\pm$ 0.3 & 97.43 $\pm$ 0.42 & 94.8 $\pm$ 0.3 & 0.67 $\pm$ 0.01   \\ \midrule
CPC          & 83.9 $\pm$ 1.5 & 83.27 $\pm$ 1.66 & 82.8 $\pm$ 1.7 & 73.94 $\pm$ 1.75 & 96.6 $\pm$ 0.4 & 94.44 $\pm$ 0.69 & 81.2 $\pm$ 0.6 & 81.14 $\pm$ 0.64 & 68.6 $\pm$ 0.5 & 0.42 $\pm$ 0.01  \\
SimCLR       & 81.0 $\pm$ 2.5 & 80.62 $\pm$ 2.31 & 78.9 $\pm$ 3.1 & 68.60 $\pm$ 2.71 & 93.0 $\pm$ 0.6 & 88.09 $\pm$ 0.97 & 93.4 $\pm$ 0.2 & 93.27 $\pm$ 0.17 & 74.6 $\pm$ 1.2 & 0.57 $\pm$ 0.08  \\
BYOL         & 89.5 $\pm$ 0.2 & 89.31 $\pm$ 0.17 & 80.1 $\pm$ 2.2 & 72.34 $\pm$ 0.60 & \textbf{98.1 $\pm$ 0.1} & \textbf{96.99 $\pm$ 0.15} & 94.9 $\pm$ 0.1 & 94.96 $\pm$ 0.07 & 73.4 $\pm$ 0.9 & 0.55 $\pm$ 0.02 \\
TNC          & 88.3 $\pm$ 0.1 & 88.28 $\pm$ 0.13 & 83.0 $\pm$ 0.9 & 73.44 $\pm$ 0.45 & 96.2 $\pm$ 0.3 & 94.47 $\pm$ 0.37 & 64.4 $\pm$ 0.3 & 64.40 $\pm$ 0.29 & 77.8 $\pm$ 0.8 & 0.55 $\pm$ 0.01  \\ 
TS-TCC       & 89.6 $\pm$ 1.0 & 90.38 $\pm$ 0.29 & \underline{83.0 $\pm$ 0.7} &  \underline{72.13 $\pm$ 1.04} & 96.9 $\pm$ 0.2 & 95.00 $\pm$ 0.24 & 97.4 $\pm$ 0.2 & 97.45 $\pm$ 0.23 & \underline{85.0 $\pm$ 1.0} & \underline{0.59 $\pm$ 0.02}  \\     
TF-C         & 88.2 $\pm$ 0.7 & 88.2 $\pm$ 0.68 & 65.8 $\pm$ 0.7 & 56.27 $\pm$ 0.24 & 96.8 $\pm$ 0.4 & 94.23 $\pm$ 0.36 & 96.7 $\pm$ 0.2 & 96.67 $\pm$ 0.21 & 78.8 $\pm$ 1.0 &  0.57 $\pm$ 0.11  \\
MHCCL        & \underline{91.6 $\pm$ 1.1} & \underline{91.77 $\pm$ 1.11} & 71.1 $\pm$ 0.4 & 61.05 $\pm$ 0.70 & 97.9 $\pm$ 0.5 & 95.44 $\pm$ 0.82 & \underline{98.7 $\pm$ 0.4} & \underline{98.71 $\pm$ 0.55} & 80.5 $\pm$ 0.5 & 0.52 $\pm$ 0.02 \\ \midrule
\textbf{\systemname{} } & \textbf{93.1 $\pm$ 0.4} & \textbf{93.08 $\pm$ 0.29} & \textbf{83.7 $\pm$ 0.3} & \textbf{74.81 $\pm$ 0.36 } & \underline{97.9 $\pm$ 0.1} & \underline{95.73 $\pm$ 0.13} & \textbf{98.8 $\pm$ 0.2} & \textbf{98.74 $\pm$ 0.34} & \textbf{87.1 $\pm$ 0.9} & \textbf{0.61 $\pm$ 0.02}  \\
    \bottomrule
  \end{tabular*}
  \label{table:results} 
\end{table*}


\subsubsection{\textbf{Baselines.}}
We compare our method to state-of-the-art (SOTA) SSL methods on time series classification tasks.
\begin{itemize}
    \item CPC~\cite{oord2019} uses autoregressive models in latent space to predict future representations, then contrasts these predictions with actual future embeddings against sampled negatives.

    \item SimCLR~\cite{chen2020simple} generates augmentation-based embeddings and optimizes the model parameters by minimizing NT-Xent loss in the embedding space. 

    \item BYOL~\cite{byol} trains two networks simultaneously: an online network predicts the target network's representation of the same data. The target network's weights are a moving average of the online network's, avoiding negative pairs.

    \item TNC~\cite{tonekaboni2021unsupervised} leverages the temporal dependencies with the assumption that close segments (neighbor) are similar while non-neighbors are dissimilar to design contrastive models.  

    \item TS-TCC~\cite{eldele2021} learns robust representation by a harder prediction task, i.e., predicting the future states of the weak time series based on the past strong augmented time series context. 

    \item TF-C~\cite{zhang2022} aims to improve the time series representation learning by augmenting the time series data through frequency perturbations, which demonstrates promising performance on relevant tasks.

    \item MHCCL~\cite{mhccl} performs a hierarchical clustering contrastive learning strategy over augmented views by introducing a two-fold strategy: an upward masking technique to update prototypes by removing outliers and a downward masking method for selecting contrastive pairs. 

\end{itemize}

\begin{table}
  \caption{\textcolor{black}{Comparative Performance on Time series classification on UEA datasets}}
  \begin{tabular}{llllll}
    \toprule
    Method & \textbf{\systemname{}}& TS2Vec & InfoTS& TS-TCC & TNC\\
    \midrule
    Avg.ACC & \textbf{0.72} &0.712 &0.714 &0.682 & 0.677\\
    Avg.Rank & \textbf{2.36}  & 2.40 &  2.633 &3.53  & 3.84 \\
  \bottomrule  
\end{tabular}
\label{tab:UEA}
\end{table}

\begin{table*}[t]
  \centering 
  \small 
  \caption{\textbf{Reduction in False Negative Pairs Performance:} This table compares the rate of false negative pairs and macro average recall (MAR) between SOTA CL (TS-TCC) and \systemname{}  across all datasets.}
  \begin{tabular*}{\textwidth}{@{\extracolsep{\fill}} l *{10}{c}}
  \toprule
              & \multicolumn{2}{c}{HAR} & \multicolumn{2}{c}{Sleep-EDF} & \multicolumn{2}{c}{Epilepsy} & \multicolumn{2}{c}{PenDigits} & \multicolumn{2}{c}{ECG Waveform} \\
    \midrule
Methods       & $FNPs \downarrow$ & $MAR \uparrow$ & $FNPs \downarrow$ & $MAR \uparrow$ & $FNPs \downarrow$ & $MAR \uparrow$ & $FNPs \downarrow$ & $MAR \uparrow$ & $FNPs \downarrow$ & $MAR \uparrow$ \\ \midrule
TS-TCC & 16.9\% & 93.1 $\pm$ 0.5 & 27.2\% & 89.6 $\pm$ 0.9 & 68.5\% & 72.1 $\pm$ 0.8 & 10.7\% & 94.2 $\pm$ 0.3 & 46.2\% & 84.2 $\pm$ 1.5 \\ 
\textbf{\systemname{}} & \textbf{3.9\%} & \textbf{95.1 $\pm$ 0.3} & \textbf{18.5\%} & \textbf{92.7 $\pm$ 0.9} & \textbf{22.4\%} & \textbf{74.1 $\pm$ 0.4} & \textbf{1.7\%} & \textbf{98.8 $\pm$ 0.3} & \textbf{26.8\%} & \textbf{86.7 $\pm$ 0.8} \\
    \bottomrule  
  \end{tabular*}
  \label{table:fnp} 
\end{table*}

\subsubsection{\textbf{Evaluation Metrics.}}
In line with previous studies, we employed widely recognized metrics for classification, namely, prediction accuracy (ACC) and macro-averaged F1 (MF1) score. Due to the severe imbalance in the ECG Waveform dataset, we employed the Area Under the Precision-Recall Curve (AUPRC) instead of MF1 for ECG, as it better manages data imbalance. 
Additionally, we incorporated Recall into our evaluation to correctly identify true positive cases, considering both true positives and false negatives. This metric effectively highlights the significance of our method in reducing false negative pairs in CL, thereby enhancing the accuracy of true positive identifications and minimizing instances of false negative predictions.
We also employ macro average recall (MAR) to provide a balanced view of all classes, especially in datasets with varying class distributions.



\begin{figure*}[t]
    \centering
    \includegraphics[width=0.95\textwidth]{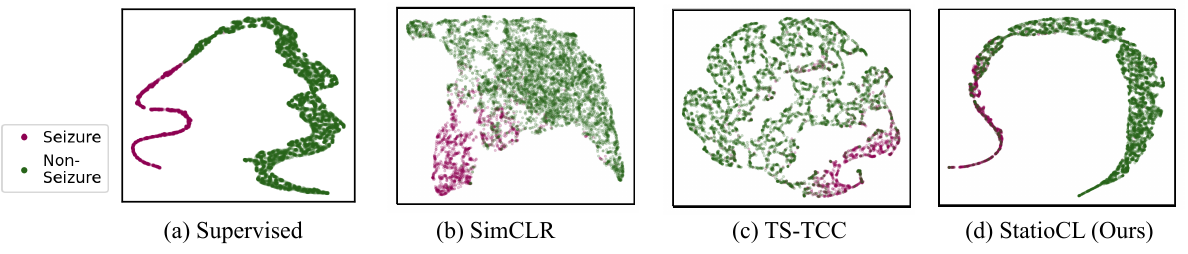}
    \caption{\textbf{Learned Representation Embedding Spaces on the Epilepsy Training Set.} }
    \label{fig:embedding}
\end{figure*}

\subsection{Time Series Classification}
\textcolor{black}{To demonstrate the effectiveness of our advanced method for time series classification, we first pre-train the encoder via CL and then freeze all its parameters for downstream tasks.}
Specifically, we follow the standard linear benchmarking evaluation scheme~\cite{chen2020simple}, where a linear classifier is trained on top of a frozen self-supervised pre-trained encoder model to assess the representations learned on various downstream datasets. 
The comprehensive results are presented in Table~\ref{table:results}.
Our proposed \systemname{} model displayed superior performance across all datasets compared to the SOTA SSL models.
In our experiments, \systemname{} demonstrates an improvement in ACC ranging from 0.3\% to 4.0\% and an enhancement in MF1 or AUPRC ranging from 0.3\% to 3.0\% across five datasets compared to the most robust baselines.
Notably, \systemname{} even surpassed the supervised learning on 4 out of 5 datasets while other SOTA SSL methods failed. This significantly underscores the efficacy of our proposed self-supervised learning framework.
\textcolor{black}{To further demonstrate the generalizability of our method to various time series modalities and classification tasks, we also conduct experiments on 30 datasets in the UEA data collections. We compare the results with baselines, including TS2Vec~\cite{ts2vec} and InfoTS~\cite{luo2023time}, with a purely unsupervised setting. The results, summarized in Table~\ref{tab:UEA}, show that \systemname{} substantially outperforms other baselines under the purely unsupervised pretraining manner, with an average accuracy of 0.72 and an average rank of 2.36.} 
These performance enhancements are a direct result of our innovative approach, which effectively reduces the occurrence of both semantic and temporal false negative pairs. By resolving these conflicts in representation learning, our method more accurately captures the underlying patterns in time series data, thereby significantly improving the efficacy of downstream classification tasks.

\begin{figure*}
\captionsetup[subfigure]{justification=centering}
\begin{subfigure}[t]{0.18\textwidth}
    \includegraphics[width=\textwidth]{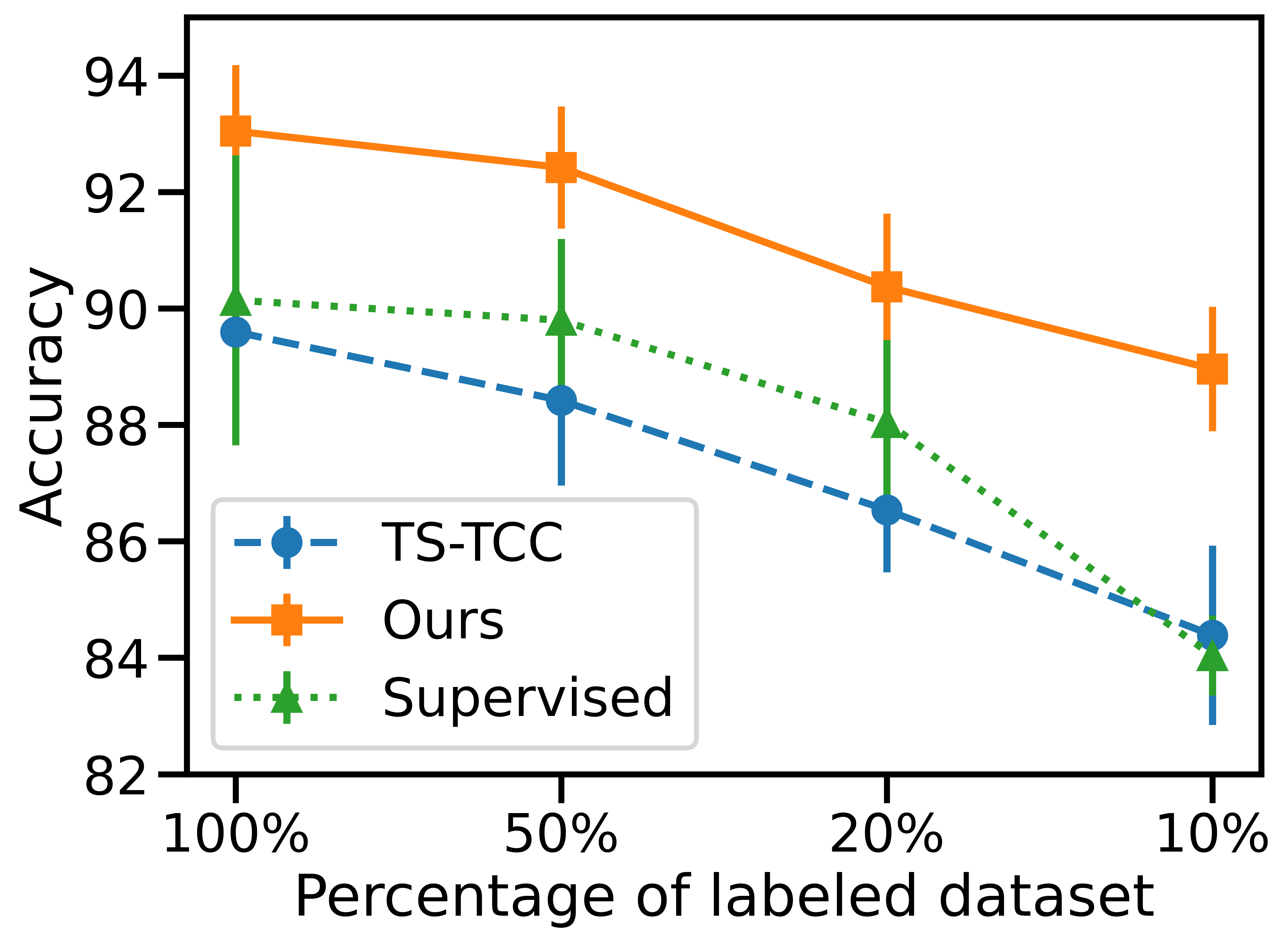}
    \caption{HAR}
\end{subfigure}
\hfill
\begin{subfigure}[t]{0.18\textwidth}
    \includegraphics[width=\linewidth]{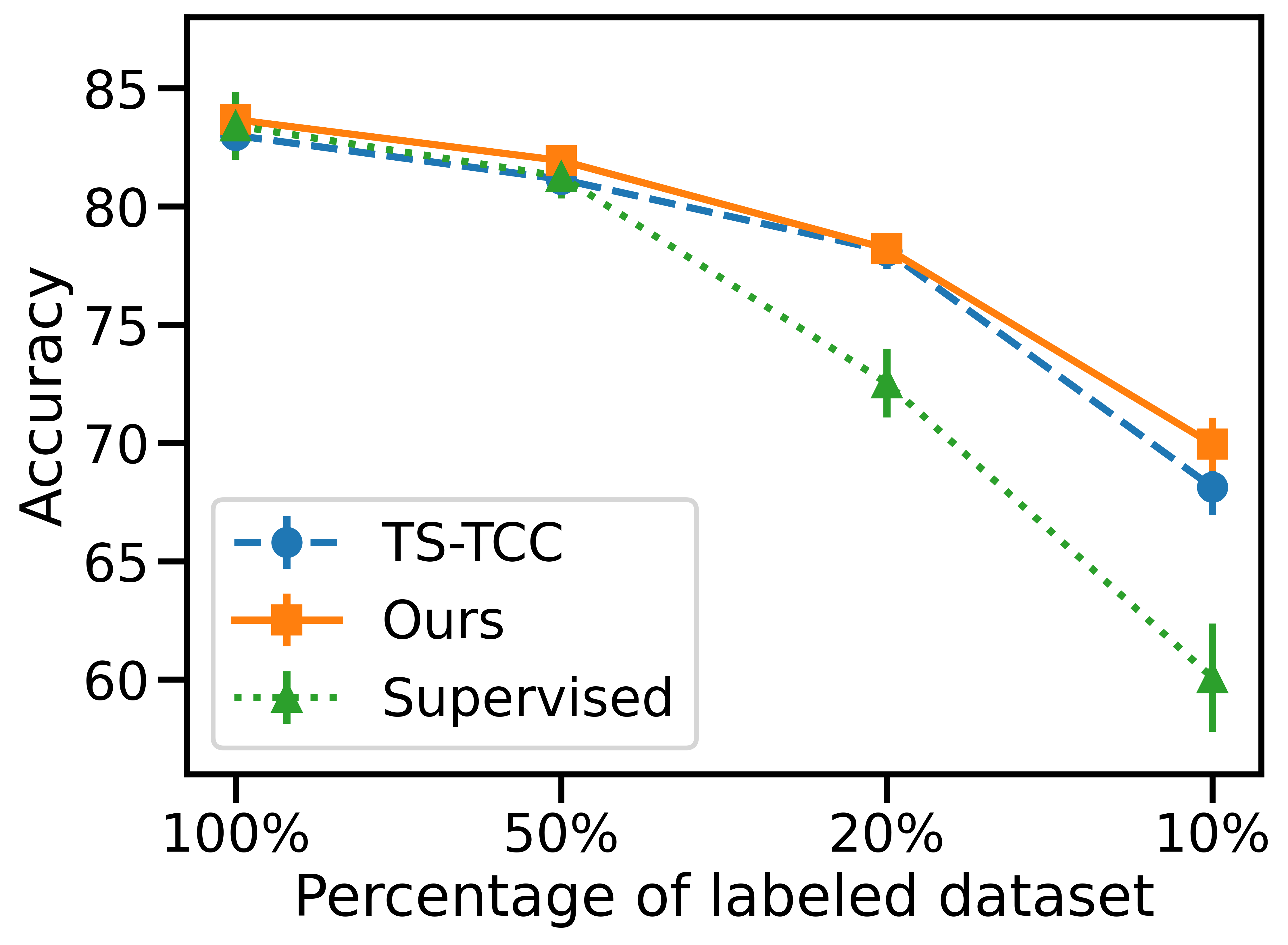}
    \caption{Sleep-EDF}
\end{subfigure}
\hfill
\begin{subfigure}[t]{0.18\textwidth}
    \includegraphics[width=\linewidth]{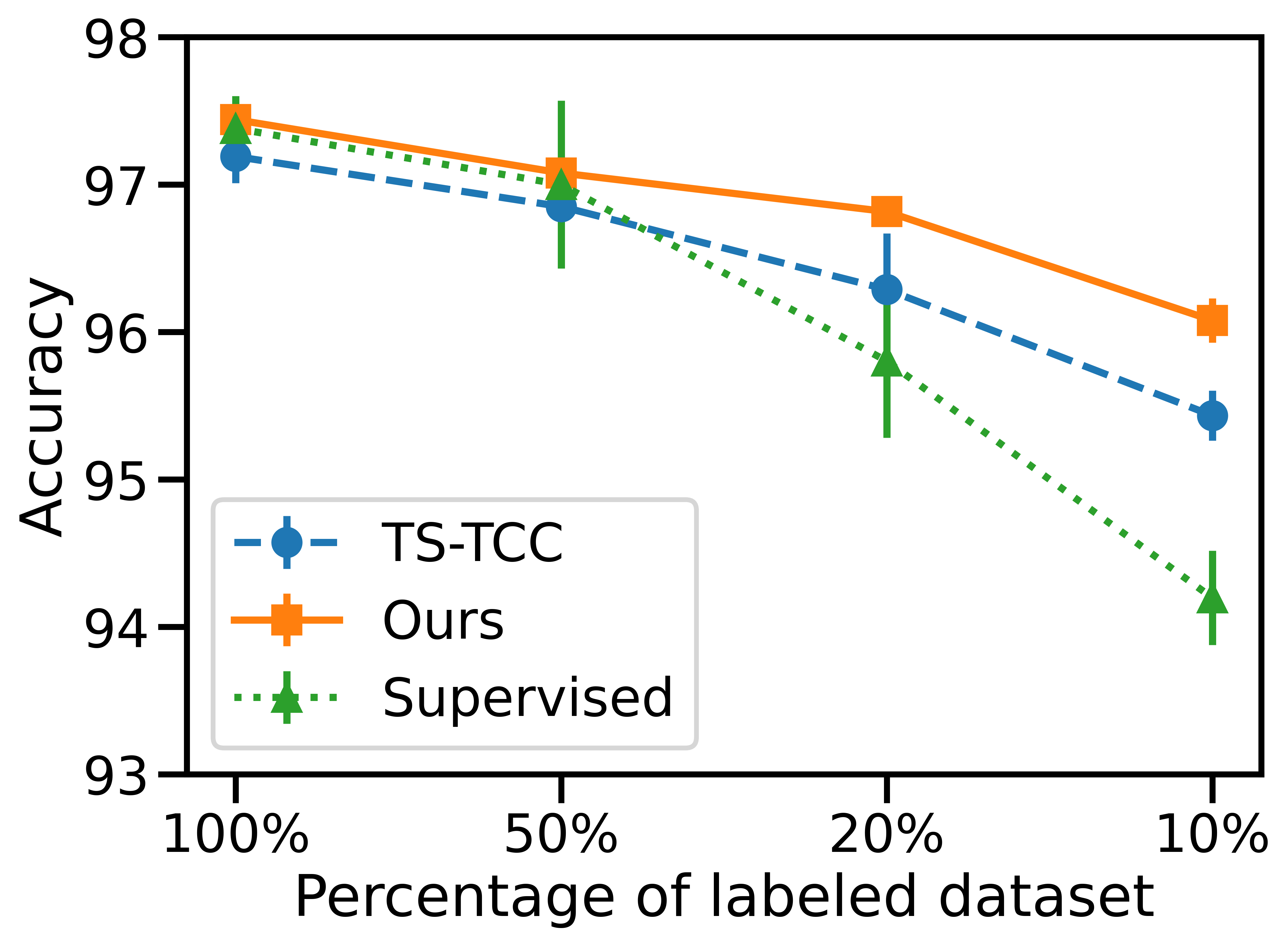}
    \caption{Epilepsy}
\end{subfigure}
\hfill
\begin{subfigure}[t]{0.18\textwidth}
    \includegraphics[width=\linewidth]{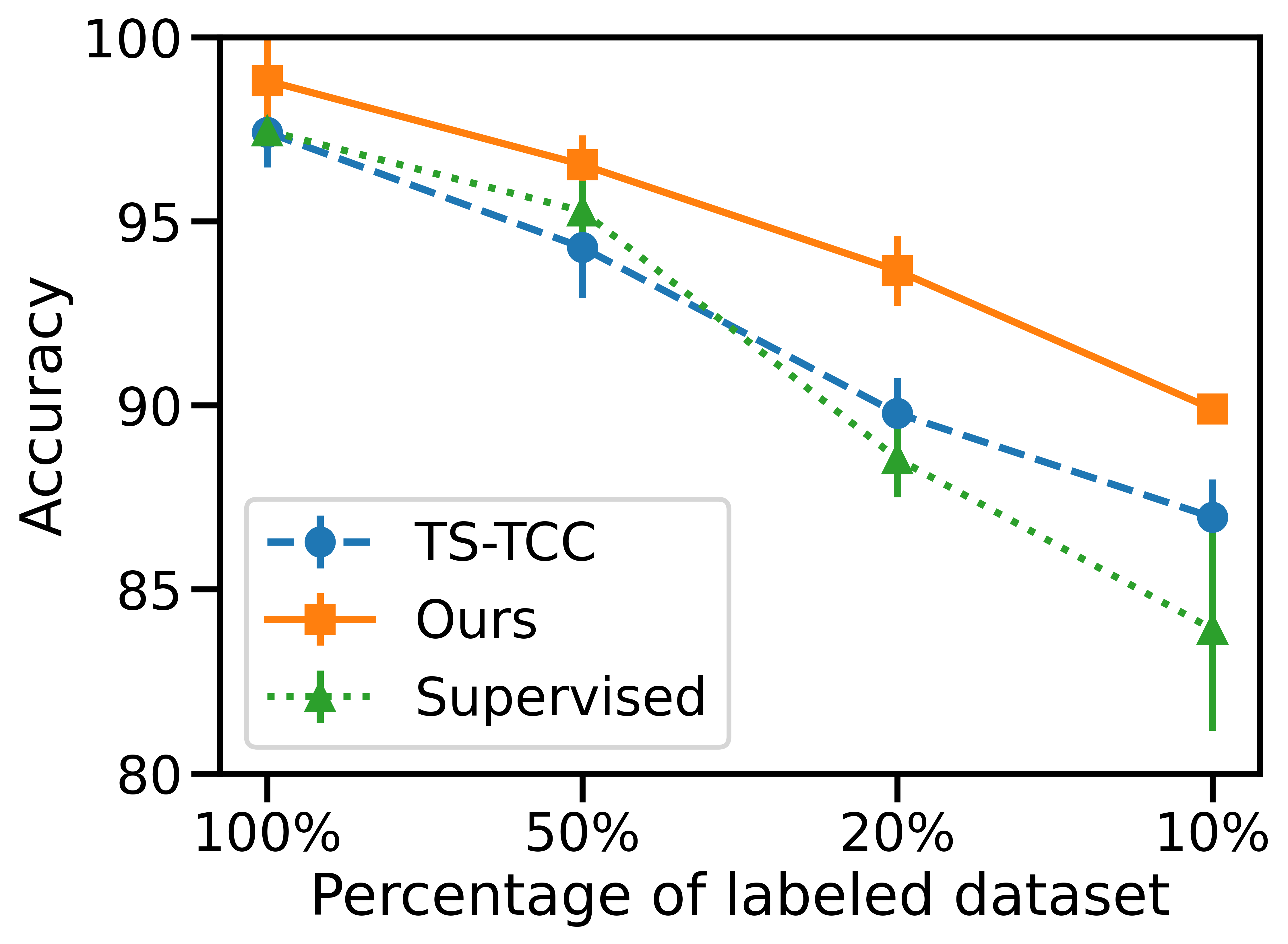}
    \caption{PenDigits}
\end{subfigure}
\hfill
\begin{subfigure}[t]{0.18\textwidth}
    \includegraphics[width=\linewidth]{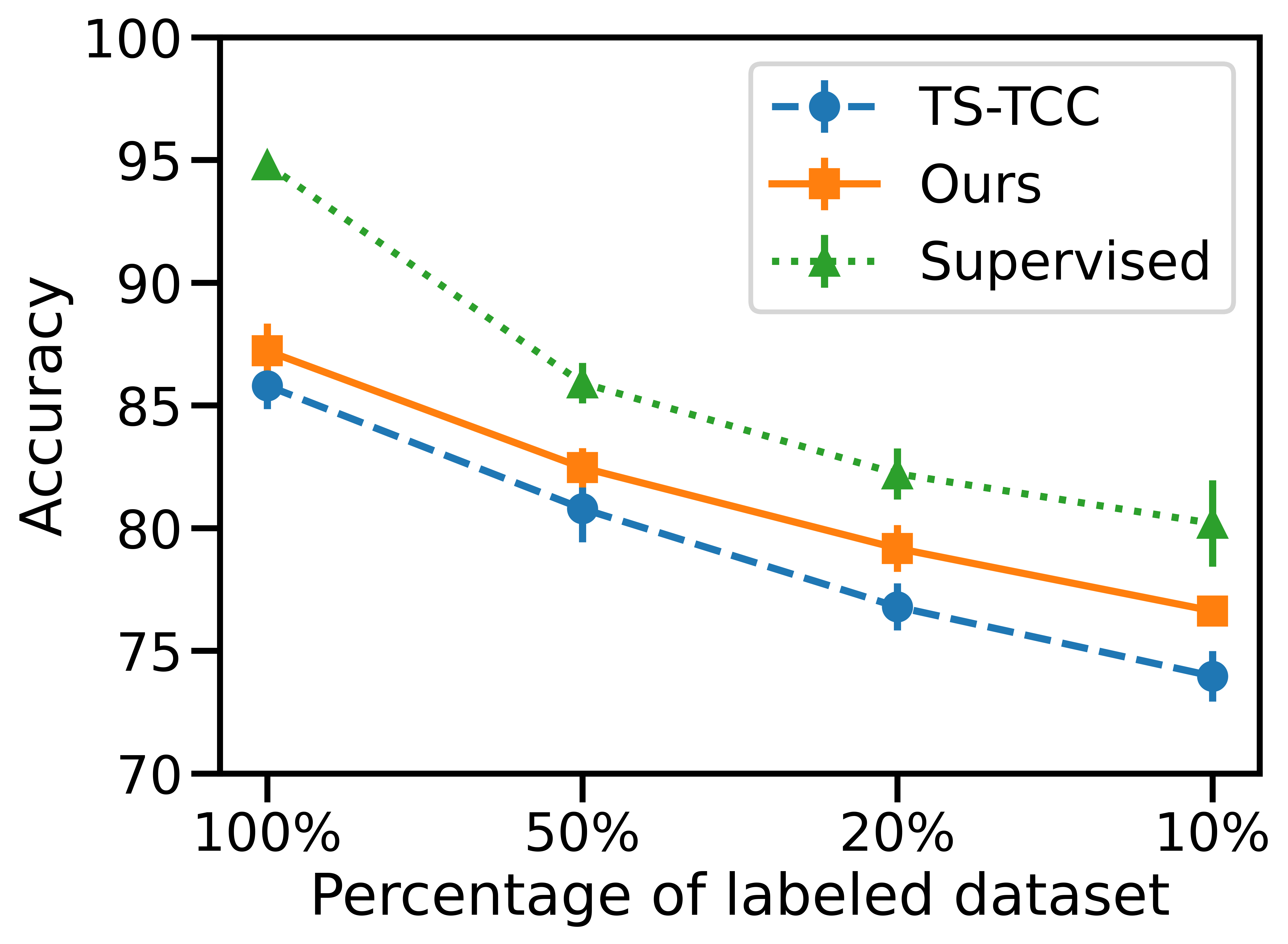}
    \caption{ECG}
\end{subfigure}
\caption{\textbf{Efficiency Analysis: This figure shows the performance of efficiency analysis of \systemname{} by varying the size of the labeled training dataset from 100\% to a sparsely labeled dataset (10\%).} 
}

\label{fig:5figs}
\end{figure*}

\subsection{False Negative Pairs Elimination Discussion}
We evaluated the effects of \systemname{} on false negative mitigation by comparing the reduction in the portion of FNPs, according to the ground-truth labels, during the training process and compared it with SOTA CL (TS-TCC). 
By effectively reducing the occurrence of FNPs, \systemname{} not only enhances the accuracy of data representation learning but is also expected to significantly enhance the model’s capability in correctly recognizing positive cases, which can be particularly evident in the Recall metric. Such capability is crucial in medical applications where the cost of missing a true positive case is high.
As detailed in Table~\ref{table:fnp}, we observed a substantial reduction in the proportion of FNPs with \systemname{}. Specifically, our framework achieved an average \textcolor{black}{19.24\%} reduction in FNPs compared to TS-TCC. This reduction directly contributed to an enhanced Recall performance. \systemname{} surpasses the SOTA baselines on an average of \textcolor{black}{2.9\%}, underscoring its effectiveness in accurately identifying true positive cases. 
Moreover, alleviating false negative pairs contributes to a more reliable embedding space. As evidenced in Figure~\ref{fig:embedding}, \systemname{} achieves better class separation than traditional contrastive learning methods and captures a similar embedding shape to the supervised method. By effectively reducing these ambiguities, \systemname{} enhances the clarity of class distinctions and potentially lowers the risk of misclassification, underscoring the method's efficacy in accurately capturing the underlying patterns of time series data.

\begin{figure*}[t]
    \centering
    \includegraphics[width=0.75\textwidth]{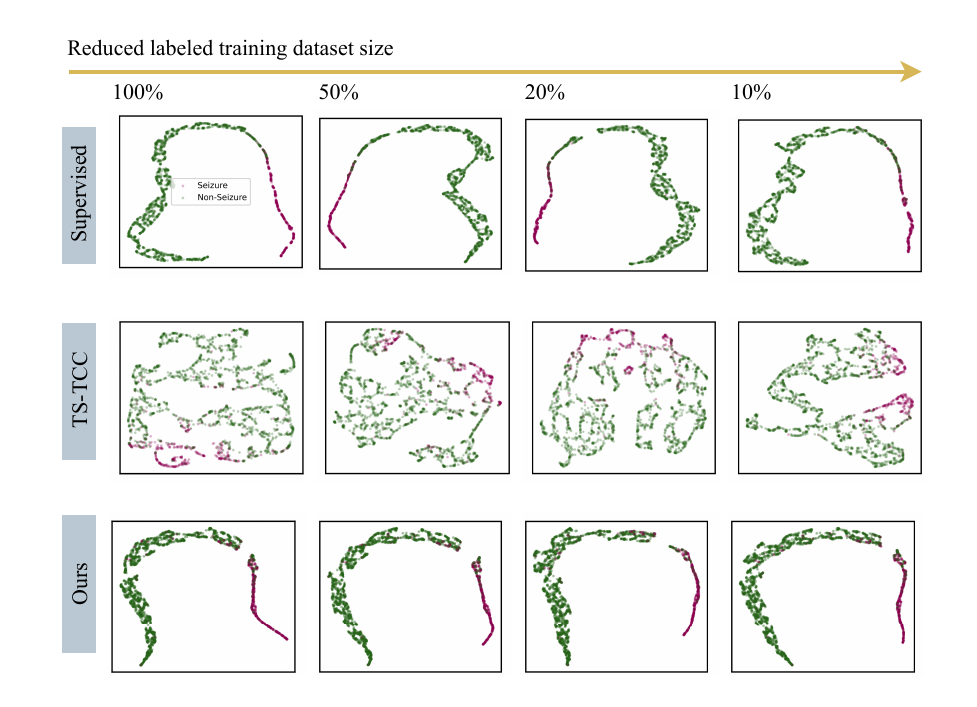}
    \caption{ Learned Representation on the Test Set of Epilepsy When Labeled Training Data Size Reduces from 100\% to 10\%.}
    \label{fig:eff_class}
\end{figure*}

\begin{table*}[t]
  \centering 
  \small 
  \caption{\textbf{Ablation Study Results of \systemname{} Across Various Time Series Datasets}.}
  \begin{tabular*}{\textwidth}{@{\extracolsep{\fill}} l * {10}{c}}
  \toprule
              & \multicolumn{2}{c}{HAR} & \multicolumn{2}{c}{Sleep-EDF} & \multicolumn{2}{c}{Epilepsy} & \multicolumn{2}{c}{\textcolor{black}{PenDigits}} & \multicolumn{2}{c}{ECG Waveform} \\
    \midrule
Methods       & ACC        & MF1        & ACC           & MF1           & ACC           & MF1       & ACC           & MF1       & ACC          & AUPRC \\ \midrule
\textbf{\systemname{}} & \textbf{93.1 $\pm$ 0.4} & \textbf{93.08 $\pm$ 0.29} & \textbf{83.7 $\pm$ 0.3} & \textbf{74.81 $\pm$ 0.36}  & \textbf{97.4 $\pm$ 0.1} & \textbf{95.73 $\pm$ 0.13} & \textbf{98.8 $\pm$ 0.2} & \textbf{98.74 $\pm$ 0.34} & \textbf{87.0 $\pm$ 0.9} &  \textbf{0.61 $\pm$ 0.02} \\
w/o TC       & 90.3 $\pm$ 0.3 & 90.29 $\pm$ 0.40 & 78.1 $\pm$ 1.2 & 70.38 $\pm$ 1.13 & 95.3 $\pm$ 3.1 & 94.85 $\pm$  1.35 & 96.7 $\pm$ 0.6 & 96.62 $\pm$  0.52 & 81.0 $\pm$ 0.3 & 0.60 $\pm$ 0.02 \\
w/o NC       & 73.3 $\pm$ 4.8 & 72.31 $\pm$ 5.24 & 64.8 $\pm$ 2.6 & 54.07 $\pm$  3.5 & 85.5 $\pm$ 3.0 & 68.77 $\pm$  2.59 & 78.9 $\pm$ 2.5 & 77.10 $\pm$ 3.19 & 85.1 $\pm$ 0.8 & 0.59 $\pm$ 0.01 \\
    \bottomrule
  \end{tabular*}
  \label{table:ablation} 

\end{table*}

\subsection{Efficiency Analysis}
We evaluated our proposed method not only on downstream datasets with fully labeled data but also in scenarios where labeled data is limited. Such limitations are common in real-world applications due to constraints on data collection resources.
By varying the size of the labeled training dataset from a full range (100\%) to a sparsely labeled dataset (10\%), we further validated the efficiency and robustness of \systemname{} in handling data scarcity. 

In Figure~\ref{fig:5figs}, we first present a performance comparison regarding Accuracy (ACC) for all five datasets when the labeled training dataset decreases from 100\% to 10\%. Notably, \systemname{} consistently outperforms the SOTA method TS-TCC across all databases, exhibiting a smaller performance drop with reduced labeled data. Specifically, TS-TCC shows an average of \textcolor{black}{9.8\%} relative decrease in performance across the datasets, while \systemname{} only experiences an \textcolor{black}{8.2\%} relative decrease. Remarkably, our model significantly surpasses supervised learning in four out of five datasets, even with only 10\% of labeled data. These results affirm the robustness and efficiency of \systemname{}, particularly in low-data regimes. 
Furthermore, Figure~\ref{fig:eff_class} reveals how \systemname{} maintains coherent latent space representations in the Epilepsy dataset, even as labeled data is reduced. In contrast, supervised methods and TS-TCC struggle to preserve distinct class discrimination under such constraints, leading to a decline in classification performance. These findings confirm the effectiveness of \systemname{} in low-data regimes and showcase its capacity to rectify the misleading embeddings caused by FNPs. 
Hence, \systemname{} directly addresses the critical challenges of contrastive learning, enhancing model efficiency and accuracy, particularly in fine tuning with less training and labeled data where FNPs could significantly distort the learning process.

\subsection{Ablation Analysis}
To evaluate each module's contribution in \systemname{}, we compare the full \systemname{} against two variants across all datasets (Table~\ref{table:ablation}).
Firstly, removing the temporal contrast (TC) module decreases performance, highlighting its importance in capturing fine-grained temporal dependencies and reducing temporal FNPs. 
Additionally, we excluded the non-stationary contrasting (NC) module, a higher decrease is observed compared to \systemname{} w/o TC among 4 out of 5 datasets. This underscores NC's role in enhancing label semantics understanding and mitigating semantic FNPs.

Interestingly, for the ECG dataset, we noticed the TC module plays a more important role than the NC module. To explore this, we analyzed the correlation between non-stationarity and label semantics by examining the ratios of stationary to non-stationary segments within each class.  Generally, a pronounced difference in these distributions among classes suggests a more evident correlation between non-stationarity and downstream task labels, which in turn augments the efficacy of the non-stationary contrast. Our examination revealed that in the ECG dataset, the ratios are fairly consistent across classes, with similar values of 5.71 and 6.35 for dominant classes 0 and 3, respectively, for class 0 and 6.35 for class 3. In contrast, the Epilepsy dataset exhibits obvious differences: 18.1 for the seizure class versus 2.1 for the non-seizure class. Such a significant difference inherently supports the non-stationarity contrast, leading to a more significant impact of this module. 
These findings demonstrate the effectiveness of the Non-Stationary and Temporal Contrast modules for accurate time series representation learning.

%% file: 04_experiments_settings.tex
To evaluate our framework's effectiveness and robustness, we selected popular benchmarks covering various domains, from sensory to general temporal data, aligning with baseline methods' experimental setups.
These datasets are as follows:
\begin{itemize}
    \item \textbf{Human Activity Recognition (HAR)}: The HAR dataset comprises data from 30 subjects, each performing six distinct activities: walking, walking upstairs, walking downstairs, standing, sitting, and lying down. The data is captured by sensors at a sampling rate of 50 Hz~\cite{har_dataset}.

    \item \textbf{Sleep Stage Classification (Sleep-EDF)}: The Sleep-EDF dataset is tailored for EEG signal classification. Each signal in this dataset is categorized into five stages:(Wake, N1, N2, N3, REM). Our analysis focused on a single EEG channel, consistent with prior research~\cite{eldele2021}.

    \item \textbf{Epileptic Seizure Prediction (Epilepsy)}: The dataset ~\cite{epi_dataset} comprises EEG recordings from 500 subjects, with each subject's brain activity recorded for 23.6 seconds. Consistent with prior research, we approached this as a binary classification task~\cite{eldele2021}. 

    \item \textbf{ECG Waveform}: The ECG dataset comprises 25 long-term Electrocardiogram (ECG) recordings of human subjects diagnosed with atrial fibrillation. Each recording spans a duration of 10 hours. The dataset captures two distinct ECG signals, both sampled at a rate of 250 Hz and annotated with four different classes\footnote{https://physionet.org/content/afdb/1.0.0/old/}.

    \item \textbf{UEA Collection}: \textcolor{black}{The UEA archive \footnote{http://www.timeseriesclassification.com/dataset.php} is one of the benchmark time series classification datasets consisting of 30 multivariate datasets with different time lengths and modalities. We evaluated \systemname{} over all datasets and reported average performance, with detailed analysis on the PenDigits dataset.}
    
\end{itemize}

We split the data into training (60\%), validation (20\%), and testing (20\%) sets following the existing setup~\cite{eldele2021, tonekaboni2021unsupervised}. More details are in Table~\ref{tab:freq}.

\begin{table}
\small
  \caption{Statistical Summary of Time Series Classification Datasets}
  \label{tab:freq}
  \begin{tabular}{lllll}
    \toprule
    Dataset & Samples  & Features & Classes & Seq. length\\
    \midrule
    HAR & 10299 & 9 & 6 & 128\\
    Sleep-EDF & 34522 & 1 & 5  & 3000 \\
    Epilepsy & 12500& 1 & 2 & 179\\
    ECG Waveform & 58766& 2 & 4 & 2500\\
    \textcolor{black}{PenDigits} & 10992 & 2 &10  & 8 \\
  \bottomrule
\end{tabular}

\end{table}

%% file: 05_conclusion.tex
\section{Conclusion and Future Work}
We introduce \systemname{}, a novel contrastive learning framework for time series representation. By comprehensively capturing the inherent complexities of time series data, non-stationarity, and fine-grained temporal dependencies, \systemname{} mitigates the issue of misleading training caused by both semantic and temporal FNPs. Our framework achieves state-of-the-art performance on real-world benchmarks, demonstrating improved data efficiency and robustness with limited labeled data. 
Therefore, our framework shows the potential for understanding intrinsic time series properties and paves the way for advanced and robust representation learning.
Future work will explore \systemname{}'s generalization to diverse tasks and its integration with advanced techniques like augmentation selection methods to develop more universal frameworks.